\pdfoutput=1

\documentclass[11pt]{article}

\usepackage[]{EMNLP2023}

\usepackage{times}
\usepackage{latexsym}

\usepackage[T1]{fontenc}

\usepackage[utf8]{inputenc}

\usepackage{microtype}

\usepackage{inconsolata}

\usepackage{enumitem}
\usepackage{refstyle}
\usepackage{xspace}
\usepackage{colortbl}

\usepackage{booktabs}
\usepackage{multirow} %
\usepackage{soul}%
\usepackage{tabularx}

\usepackage{amsmath}
\usepackage{pifont}

\usepackage{graphicx}

\usepackage{cleveref}
\crefformat{section}{\S#2#1#3}
\crefformat{subsection}{\S#2#1#3}
\crefformat{subsubsection}{\S#2#1#3}
\crefrangeformat{section}{\S#3#1#4 to~\S#5#2#6}
\crefmultiformat{section}{\S#2#1#3}{ and~\S#2#1#3}{, #2#1#3}{ and~#2#1#3}
\Crefformat{figure}{#2Fig.~#1#3}
\Crefmultiformat{figure}{Figs.~#2#1#3}{ and~#2#1#3}{, #2#1#3}{ and~#2#1#3}
\Crefformat{table}{#2Tab.~#1#3}
\Crefmultiformat{table}{Tabs.~#2#1#3}{ and~#2#1#3}{, #2#1#3}{ and~#2#1#3}
\Crefformat{appendix}{#2Appx.~\S#1#3}
\crefformat{algorithm}{Alg.~#2#1#3}
\Crefformat{equation}{#2Eq.~#1#3}

\newcommand{\stitle}[1]{\vspace{1ex} \noindent{\bf #1.}}

\usepackage{caption}
\usepackage{subcaption}
\usepackage{tcolorbox}
\usepackage{setspace}

\usepackage{tcolorbox}

\title{A Causal View of Entity Bias in (Large) Language Models}

\author{
Fei Wang$^\dagger$ \; Wenjie Mo$^\dagger$ \; Yiwei Wang$^\ddagger$ \; Wenxuan Zhou$^\dagger$  \; Muhao Chen$^{\dagger\sharp}$ \\
$^\dagger$University of Southern California; 
$^\ddagger$University of California, Los Angeles; \\
$^\sharp$University of California, Davis \\
\texttt{\{fwang598,jackymo,zhouwenx\}@usc.edu; wangyw.evan@gmail.com;} \\
\texttt{muhchen@ucdavis.edu}
}

\begin{document}
\maketitle

\begin{abstract}
    Entity bias widely affects pretrained (large) language models, causing them to rely on (biased) parametric knowledge to make unfaithful predictions. Although causality-inspired methods have shown great potential to mitigate entity bias, it is hard to precisely estimate the parameters of underlying causal models in practice. The rise of black-box LLMs also makes the situation even worse, because of their inaccessible parameters and uncalibrated logits. To address these problems, we propose a specific structured causal model (SCM) whose parameters are comparatively easier to estimate. Building upon this SCM, we propose causal intervention techniques to mitigate entity bias for both white-box and black-box settings. The proposed causal intervention perturbs the original entity with neighboring entities. This intervention reduces specific biasing information pertaining to the original entity while still preserving sufficient semantic information from similar entities. Under the white-box setting, our training-time intervention improves OOD performance of PLMs on relation extraction (RE) and machine reading comprehension (MRC) by 5.7 points and by 9.1 points, respectively. Under the black-box setting, our in-context intervention effectively reduces the entity-based knowledge conflicts of GPT-3.5, achieving up to 20.5 points of improvement of exact match accuracy on MRC and up to 17.6 points of reduction in memorization ratio on RE.\footnote{Our code is available at \url{https://github.com/luka-group/Causal-View-of-Entity-Bias}}
\end{abstract}
\section{Introduction}

\begin{figure}[t]
\centering
\includegraphics[width=\columnwidth]{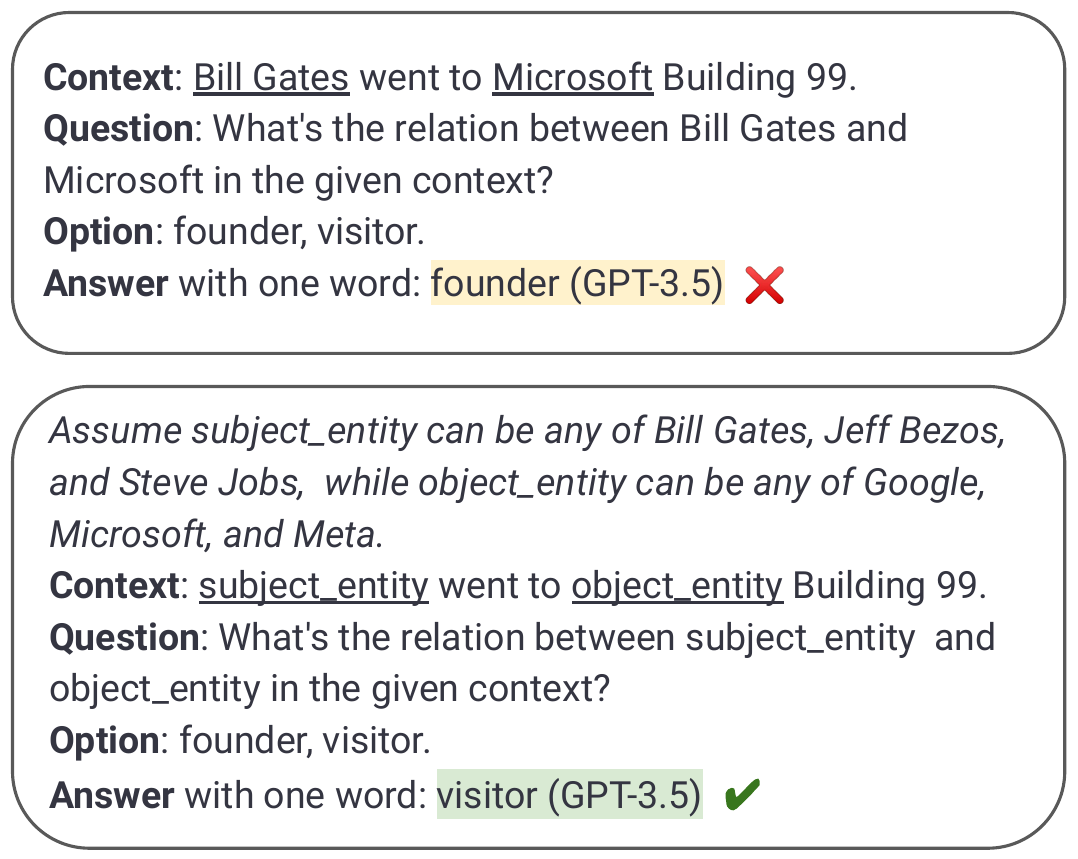}
\caption{An example of entity bias in GPT-3.5. Our in-context intervention mitigates the conflicts between parametric knowledge and contextual knowledge. 
}
\label{fig/example}
\end{figure}

Entity bias  \cite{longpre2021entity,wang2022should,xu2022does,peng2020learning,qian-etal-2021-annotation,hermann2015teaching} refers to an undesirable phenomenon where models overly rely on %
prediction shortcuts triggered by specific entities to make spurious predictions. 
For example, given the sentence \textit{``Bill Gates went to Microsoft Building 99,''} models may be misled by their memory of the entities \textit{Bill Gates} and \textit{Microsoft}, saying the relation between them in this context is \textit{founder} rather than \textit{visitor}, as shown in \Cref{fig/example}.
Recent studies show that entity bias widely %
affects pretrained (large) language models (LLMs; \citealt{longpre2021entity,yan2022robustness,zhou2023context}).
These models have a tendency to disregard contextual information that contradicts or is infrequently reported in the pretrained corpus, while excessively relying on (biased) parametric knowledge \cite{longpre2021entity} to make unfaithful predictions and perpetuate bias.

Prior studies have proposed multiple causality-inspired methods to mitigate entity bias %
\cite{zhang2017position,nan2021uncovering,wang2022should,zhu2022generalizing}.\footnote{Although \citet{zhang2017position} do not mention causal theory, the proposed entity masking does follow a relevant principle to cut off causal links between specific entities and labels.}
Despite their potential, the causal models underlying these methods are flawed in practice, primarily because of imprecise parameter estimation.
For example, some causal models necessitate estimating the probability distribution over labels when given a sentence that is devoid of entities or contextual information \cite{zhang2017position,wang2022should}.
These methods either lose predictive information about entities,
or are prone to erroneous representation without contextualization.
The other critical problem is the difficulty of applying these methods to black-box LLMs, of which parameters are inaccessible and logits are uncalibrated.

\begin{figure}[t]
\centering
\includegraphics[width=\columnwidth]{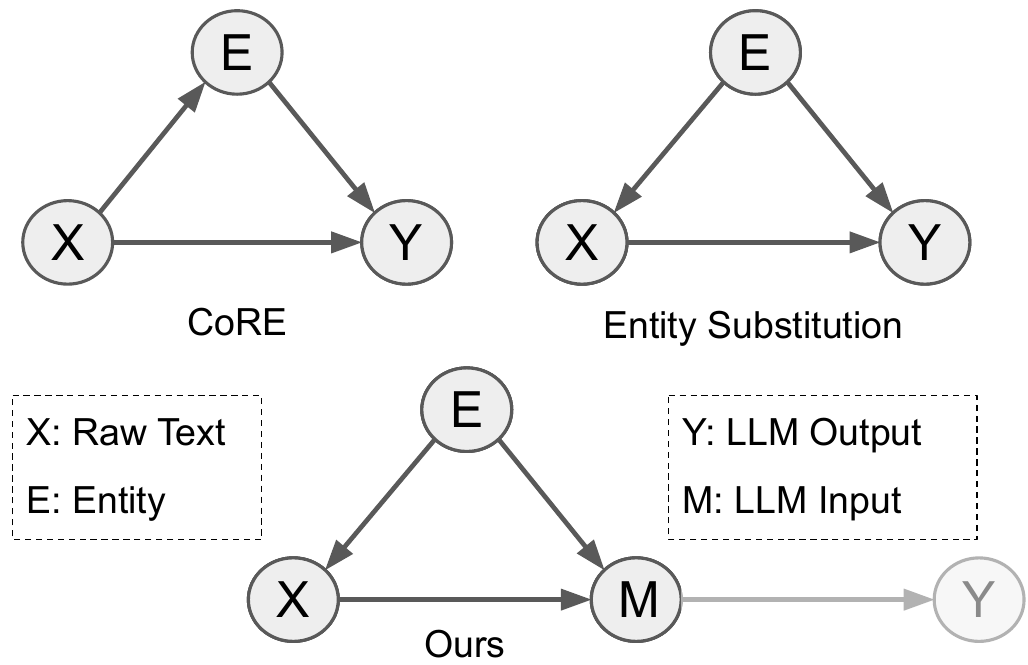}
\caption{Structured causal models revealing entity bias.}
\label{fig/scm}
\end{figure}

To address the aforementioned problems, the \emph{first} contribution of this paper is a \textbf{causal analysis of entity bias mitigation methods} (\Cref{sec/scm}).
We examine and compare the structured causal models (SCMs) behind existing methods.
We find that, among the theoretically equivalent causal models \cite{verma1990equivalence}, there exists a specific SCM whose parameters are comparatively easier to estimate. 
As shown in \Cref{fig/scm}, the proposed SCM only requires to intervene input entities to mitigate the presence of spurious features before passing them to the subsequent neural layers.
Moreover, it retains the entity type information\footnote{Entity type information plays a crucial role in entity-driven tasks. For example, without knowing a more specific location type, it is impossible to differentiate between relations \textit{born\_in\_city} and \textit{born\_in\_country}.} at an appropriate level of granularity without requiring explicit entity typing.

The \emph{second} contribution of this paper is a \textbf{training-time causal intervention technique} for mitigating entity bias based on the proposed SCM (\Cref{sec/white-box}).  
Specifically, we identify entities that are likely to share similar predictive information with the given entity.
During training, we perturb embedding of the given entity within a convex hull constructed by embeddings of similar entities.
During inference, we represent the entity with the center of the convex hull.
Taking advantage of the continuous nature of the embedding space, this intervention does not rely on models specifically trained on natural language to estimate the label distribution of unnatural text, nor does it sacrifice predictive entity or contextual information.

The \emph{third} contribution of this paper is to transform the training-time intervention into \textbf{in-context intervention for black-box LLMs} whose parameters are inaccessible, and logits are uncalibrated (\Cref{sec/black-box}). 
A significant advantage of the proposed SCM is that the causal intervention is carried out at the input layer, enabling its implementation within an in-context setting.
Specifically, we replace entities with placeholders and define each placeholder by examples -- a set of similar entities.
For example, we can replace \textit{Bill Gates} in \Cref{fig/example} with \textit{subject\_entity} and prepend the prompt, \textit{``Assume that subject\_entity can be any of Steve Jobs, Bill Gates, and Jeff Bezos"}, to the input. 
This in-context intervention can be applied to any black-box LLM without additional cost.

Experiments on relation extraction (RE) and machine reading comprehension (MRC) show that the proposed causal intervention techniques are effective for both white-box and black-box LLMs. 
Under the white-box setting (\Cref{sec/exp/white}), our training-time intervention significantly improves out-of-distribution performance of RoBERTa \cite{liu2019roberta} on RE by 5.7 points and SpanBERT \cite{joshi2020spanbert} on MRC by 9.1 points, comparing with the vanilla version.
Under the black-box setting (\Cref{sec/exp/black}), our in-context intervention effectively reduces the entity-based knowledge conflicts \cite{longpre2021entity} and improves the task performance of GPT-3.5.\footnote{https://platform.openai.com/docs/models/gpt-3-5} Specifically, our method outperforms the best baseline by up to 20.5 points of exact match accuracy on MRC and reduces the memorization ratio by up to 17.6 points on RE.
Further analyses reveal the crucial role of the number of neighboring entities $k$ in balancing the predictive information and biasing information from entities, and the necessity of entity placeholder definition for in-context intervention.

\section{Related Work}

\stitle{Entity Bias in LLMs}
LLMs memorize factual knowledge in their parameters during pretraining~\cite{roberts-etal-2020-much,jiang-etal-2020-know} and show promising results in answering factual questions~\cite{petroni-etal-2019-language,brown2020language,weifinetuned}.
However, the parametric knowledge may be inaccurate due to the misinformation in the training corpus~\cite{lin-etal-2022-truthfulqa} or outdated as the world evolves~\cite{liska2022streamingqa,kasai2022realtime}.
In such scenarios, it is critical for LLMs to update their predictions when provided with contextual evidence.
However, previous studies~\cite{longpre2021entity,qian-etal-2021-annotation,yan2022robustness} observe that language models may take entities as shortcuts, leading to spurious predictions based solely on parametric knowledge.
This bias becomes more prominent when the evidence contains infrequent or conflicting knowledge compared to the training corpus.

To mitigate this bias, previous work~\cite{longpre2021entity,chen-etal-2022-rich,li2022large,zhou2023context} introduces the entity substitution technique, which involves constructing counterfactual data by randomly replacing the entities, and updating the language models either by finetuning or in-context learning.
Although showing improved results, these techniques are empirical and lack theoretical backgrounds.
In this paper, we theoretically analyze the entity bias problem from a causal view.
Furthermore, we propose a causal intervention method that surpasses the performance of entity substitution.

\stitle{Debiasing with Causal Intervention} %
LLMs have been revealed with bias problems, for which
literature has paid much attention in order to mitigate their adverse effects
\cite{sweeney2019transparent,zhang2020demographics,venkit2021identification,lalor2022benchmarking}.
Recent debiasing techniques incorporate  the concept of counterfactual inference, and have been applied %
in various tasks for bias mitigation \cite{niu2021introspective,qian2021counterfactual,wang2022should}.
One dominant technique is based on causal mediation analysis \cite{udomcharoenchaikit2022mitigating}, 
which involves decomposing the total effect into pure direct effect and total indirect effect. 
In this context,
\citet{wang2022should} utilize total direct effect and total effect to debias the relation extraction.
Apart from debiasing, causal mediation analysis can be used to analyze biases in LLMs \cite{vig2020causal,finlayson2021causal}.

In addition to intervening causal mediator, previous studies have also explored %
confounder analysis
\cite{keith2020text,qian2021counterfactual,feder2022causal,weld2022adjusting}.
A confounder is a variable that influences both the input and the output, causing a spurious correlation between them.
Typically, the de-confounder process applies the \textit{do}-calculus \cite{pearl2012calculus} to compute the prediction assuming that the value of the confounder variable is not the observed one but follows its natural distribution \cite{zhang2020causal,tian2022debiasing}.
Our approach is also based on confounder analysis. While nearly all the aforementioned approaches request a white-box accessibility of the model with at least logits of predictions, this work represents a pilot study of deconfounder method that applies to purely black-box LLMs.
\begin{figure*}[t]
\centering
\includegraphics[width=\textwidth]{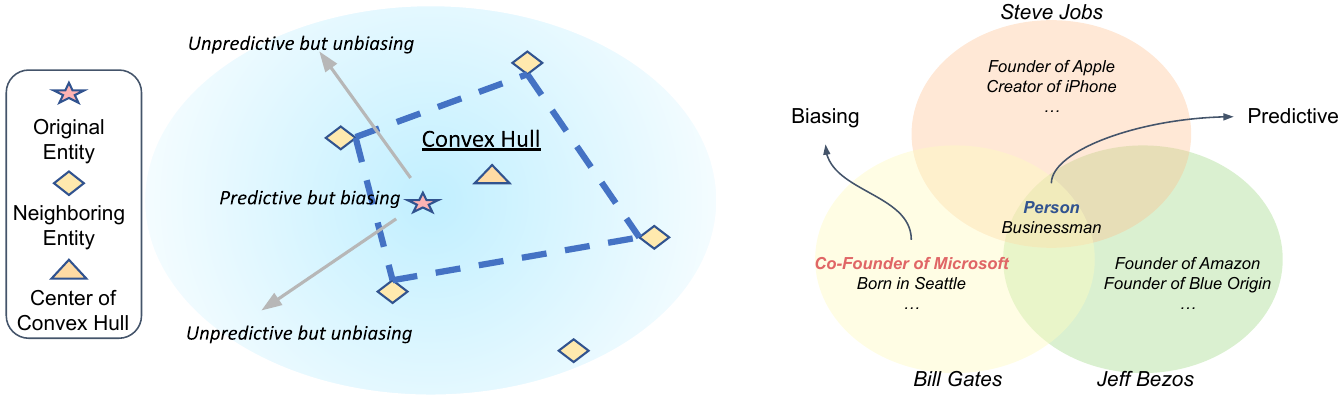}
\caption{Left: Training-time intervention with $k=4$. Right: Example of predictive and biasing information.}
\label{fig/training_time}
\end{figure*}

\section{Method}
In this section, we first analyze methods for mitigating entity bias in a causal view and propose an easy-to-estimate SCM as a theoretical basis (\Cref{sec/scm}). Based on the proposed SCM, we design a training-time intervention technique for white-box LLMs (\Cref{sec/white-box}) and an in-context intervention technique for black-box LLMs (\Cref{sec/black-box}).

\subsection{Causal Analysis of Entity Bias}
\label{sec/scm}

To compare existing methods in the same context, we analyze the structured causal models (SCMs) behind them.
\Cref{fig/scm} shows two typical SCMs for entity bias mitigation methods, where $X$ refers to the raw input, $E$ refers to entities, and $Y$ refers to the label. 
The links $X \rightarrow Y \leftarrow E$ show that LLMs rely on both predictive information from the whole input and the biasing information from specific entities to make the prediction.
The links $E \rightarrow X$ and $X \rightarrow E$ assume that the context is written down with the entity in mind or vice versa.
As discussed by \citet{verma1990equivalence}, we cannot differentiate between these two directions merely based on statistical observations. 
Indeed, the two SCMs with opposite links between $X$ and $E$ are equivalent according to the Bayes' theorem:
\begin{align}    
    &P(X)P(E|X)P(Y|X,E) \nonumber \\
    = & P(Y,X,E) \nonumber \\
    = & P(E)P(X|E)P(Y|X,E) \nonumber 
\end{align}

As revealed by these SCMs, entity bias exists in LLMs because entities serve as either confounders or mediators.
Thus, the bias can be mitigated through causal intervention, such as backdoor adjustment 
$$P(Y|do(X)) = \sum_E P(Y|X,E)P(E),$$ 
which eliminates the influence of a specific variable (in this context, $E$) by assigning values to this variable.
However, previous SCM-based debiasing methods exhibit divergent performances, since they estimate different (conditional) probabilities using different surrogates when performing the causal intervention.
For example, counterfactual analysis by \citet{wang2022should} estimates and deducts the biasing effect of entities on labels by masking the context,
while \citet{zhang2017position} and \citet{longpre2021entity} directly remove the effect of entities by entity masking or substitution.
None of them estimates the causal effects of entity names precisely, due to the highly complex architectures of LLMs, which account for their unsatisfactory performance on mitigating entity bias.

In this work, we consider the SCM in \Cref{fig/scm}, whose parameters are much easier to estimate in practice.
Since most LLMs follow a sequential structure by stacking neural layers, mitigating the entity bias in one layer %
will also mitigate the entity bias in subsequent layers. 
The underlying logic is simple -- if we block the spurious features in the input, there will be no spurious correlations to capture.
Therefore, we propose to mitigate the entity bias in the input layer $M$, which could be an embedding layer or a prompt layer.
Obviously, $P(M|X,E)$ can be estimated more accurately and efficiently than $P(Y|X,E)$, because there is no need to run the whole model, ensuring less error propagation and computational cost.
To further improve the estimation by retaining as much predictive information as possible, we propose to estimate $P(M|do(X))$ by perturbing the entity with similar entities rather than masking it.
In the following sections, we will show how to realize the proposed causal intervention on both white-box and black-box LLMs.

\begin{figure*}[t]
\centering
\includegraphics[width=\textwidth]{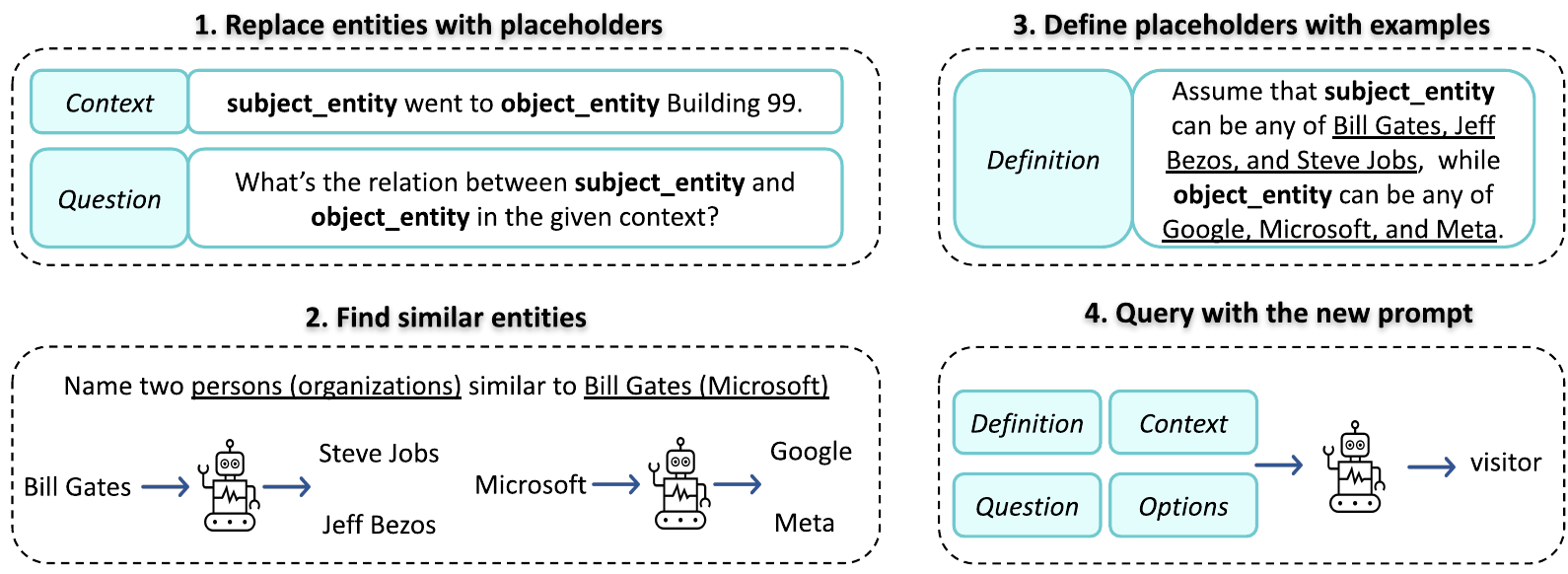}
\caption{In-context intervention for black-box LLMs. We take relation extraction as an example.}
\label{fig/llm}
\end{figure*}

\subsection{Training-time Intervention}
\label{sec/white-box}

For white-box models of which  the parameters are accessible, we can effectively address their internal bias through training-time intervention.
In the case of entity bias identified by the proposed SCM, we realize the causal intervention by perturbing the input entities or entity tokens using their neighboring counterparts in the embedding space, as shown in \Cref{fig/training_time} (Left). 
For each entity presented in the input text, we first find its top $k$ nearest neighbors according to embedding distance.
Then we construct the smallest convex hull\footnote{This convex hull-bounded perturbation is inspired by \citet{dong2021towards}, where perturbation within a convex hull formed by synonyms is used to improve model robustness against word substitutions.} to cover the original entity and neighboring entities.
Due to the continuous nature of the embedding space, the embeddings within the convex hull approximately represent the same predictive information as a whole.
The entity-specific biasing information, which has the potential to trigger spurious shortcuts, gradually diminishes from the original entity towards the border of the convex hull.

During training, we introduce perturbations to the entity embedding by replacing it with a random embedding selected from within the convex hull. In this way, the convex hull bounded the predictive information, while random sampling further introduces noises and increases the diversity of data for robust training.
During inference, we replace the original entity embedding with the center of the convex hull, in order to balance the trade-off between predictive and biasing information. 
\Cref{fig/training_time} (Right) provides an example of the information preserved through such intervention.
By replacing the entity \textit{Bill Gates} with the center of the convex hull, encompassed by its neighboring entities, such as \textit{Steve Jobs} and \textit{Jeff Bezos}, we effectively retain the shared predictive information (e.g., person), while mitigating the biasing information (e.g., founder of Microsoft).
That is to say, the convex hull-bounded perturbation serves as an effective estimation of $P(M|do(X))$.

\subsection{In-context Intervention}
\label{sec/black-box}
The rise of Web services powered by black-box LLMs, such as GPT-3.5, introduces new challenges for mitigating entity bias, demanding debiasing methods that do not require accessible model weights and prediction logits.
As discussed in \Cref{sec/scm}, a key advantage of our SCM is that the deconfounder operation is merely on the input layer. 
In the context of black-box LLMs, the input is the user-provided prompt.
Thus, we perform the causal intervention solely through modifying prompts to resolve entity bias.
We propose a four-step (test-time) in-context intervention technique for black-box LLMs.  \Cref{fig/llm} shows the whole process. 

First, we replace the original entity mention in the input with abstract placeholders (e.g., \texttt{[ENTITY]}). This step effectively mitigates any biasing information from the original entity names, because the placeholders are semantic-neutral. However, this step also eliminates predictive information from entities. We show in \Cref{sec/black/ablation} that, without proper definition for the placeholder, models can easily fail to answer questions. In the next two steps, we construct definitions to provide predictive information for each placeholder while introducing minimal additional biasing information.
Second, we query the LLM to name $k$ entities similar to the original one (e.g., $E_o$).\footnote{Here, we rely on the entity knowledge possessed by LLMs. However, it is possible to replace the LLM with external databases or tools in this step.} These generated entities (e.g., $E_a$ and $E_b$) present similar predictive information as the original entity, and are able to fulfill the same function as neighboring entities in \Cref{sec/white-box}.
Third, we define the placeholder with the original entity and generated entities. For example, we can verbalize the definition as \textit{``Assume \texttt{[ENTITY]} can be any of $E_o$, $E_a$ and $E_b$''}. This definition encourages the LLM to find common properties of given entities rather than relying on biasing information of one specific entity. The resulting placeholder along with its definition serves as an effective estimation of $P(M|do(X))$.
Finally, we prepend the placeholder definition to the modified context and question, and query the LLM with the new prompt. 
This four-step adjustment ensures that the resulting prompt is free of specific biasing information pertaining to the original entity while still preserving sufficient predictive information by considering given entity examples as a whole.

\begin{table*}[t]
\centering
\small
\begin{tabular}{lccc|ccc} \toprule
& \multicolumn{3}{c|}{RE (F1)} & \multicolumn{3}{c}{MRC (EM)} \\ \cmidrule{2-4} \cmidrule{5-7}
                               & ID & OOD & $\Delta$ & ID & OOD & $\Delta$ \\  \midrule
Vanilla Model & 71.1$_{\pm 0.9}$  & 62.3$_{\pm 0.6}$ & $-12.4\%$ & 79.1$_{\pm 0.1}^\dagger$ & 63.1$_{\pm 0.8}^\dagger$ &  $-20.2\%$       \\ \midrule
\rowcolor{gray!20} + Continual Pretraining \cite{yan2022robustness}$^*$ & - & - & - & \textbf{79.6}$_{\pm 0.6}^\dagger$ & 65.9$_{\pm 1.1}^\dagger$ &  $-17.2\%$ \\\midrule
+ CoRE    \cite{wang2022should}  & \textbf{71.3}$_{\pm 0.3}$ &  61.2$_{\pm 0.6}$ & $-14.2\%$ & - & - & -   \\
+ Entity Mask    \cite{zhang2017position} & 61.4$_{\pm 0.5}$ & 61.9$_{\pm 0.5}$ & \boldmath{$+0.9\%$} & 75.7$_{\pm 0.6}$ & 62.9$_{\pm 0.4}$ & $-16.9\%$  \\
+ Entity Substitution \cite{longpre2021entity} & 66.6$_{\pm 0.6}$ & 65.8$_{\pm 0.3}$  & $-1.2\%$ & 76.4$_{\pm 0.8}$  &  70.8$_{\pm 1.5}$ &  $-7.3\%$          \\ \midrule
+ Ours  & 70.8$_{\pm 0.3}$ & \textbf{68.0}$_{\pm 0.3}$ & $-3.9\%$ & 77.0$_{\pm 0.7}$ & \textbf{72.2}$_{\pm 0.5}$  & \boldmath{$-6.2\%$}     \\ \bottomrule
\end{tabular}
\vspace{-0.5em}
\caption{Results under white-box setting. We report the average F1/EM score and standard deviation of three runs. $\Delta$~shows the relative performance change between ID and OOD. The best number of each column is in bold. $^*$~Continual pretraining is not directly comparable to finetuning methods. $^\dagger$~Numbers copied from \citet{yan2022robustness}. }
\label{tab/roberta}
\end{table*}

\section{White-Box Experiments}
\label{sec/exp/white}

In this section, we evaluate our training-time intervention under the white-box setting.

\subsection{Experimental Setup}

\stitle{Datasets and Metrics}
We evaluate our methods on relation extraction (RE) and machine reading comprehension (MRC). For both tasks, we finetune models on an in-distribution (ID) training set and evaluate models on both ID and out-of-distribution (OOD) test sets.
For RE, we adopt TACRED \cite{zhang2017position} as the ID dataset and EntRED \cite{wang2023entred} as the OOD dataset, and report micro-F1 score. In both datasets, entities in each sentence are given.
For MRC, we adopt TriviaQA \cite{joshi2017triviaqa} as the ID dataset and its answer-substituted version \cite{yan2022robustness} as the OOD dataset, and report exact match (EM) score. Following \citet{yan2022robustness}, we hold out 10\% of the training data for development and evaluate models on the original development set. We use the DBName version of their OOD dataset.
For all metrics, we report the average score with standard deviation of three runs.

\stitle{Baselines}
We compare our methods with the following baselines.
\textit{Entity Mask} \cite{zhang2017position} masks the subject and object entities in the sentence with special tokens.
\textit{Entity Substitution} \cite{longpre2021entity} randomly selects an entity of the same type to substitute the original entity.
\textit{CoRE} \cite{wang2022should} applies counterfactual inference by computing the difference between the prediction made with the entire sentence and the prediction made with only the entities observed.
\textit{Continual Pretraining} \cite{yan2022robustness} introduces an intermediate pretraining stage to the backbone model with the objective of recovering masked entities.

\stitle{Implementation Details}
For RE, we apply RoBERTa \cite{liu2019roberta} as the backbone model following previous works \cite{zhou2022improved,wang2022should}. 
We use the \texttt{entity\_marker\_punct} input format from \citet{zhou2022improved} in main experiments, in order to mitigate the impact of explicit entity type information on our analysis of entity bias.   %
For MRC, we apply SpanBERT \cite{joshi2020spanbert} as the backbone model following \citet{yan2022robustness}. 
Since entities are not given in MRC datasets, we use the same named entity recognition tool used by \citeauthor{yan2022robustness} to extract entities. Since the detected entities could be noisy and incomplete, we perform our method upon answer-substituted training set ensuring all answer entities are perturbed as strong as \textit{Entity Substitution}.
Since RoBERTa and SpanBERT lack entity-level embeddings, we apply our causal intervention to each token embedding within the entity mention instead.
To construct convex hull, We select neighboring tokens based on their Euclidean distance to the original token in the embedding space.
For both tasks, we perform training-time intervention on each entity token with $k=3$.
While further data augmentation is always possible, for a fair comparison, we finetune all the models with the same amount of data.
More implementation details are in \Cref{sec/app/white}.

\subsection{Results}
As shown in \Cref{tab/roberta}, the vanilla RoBERTa and SpanBERT experiences significant declines in performance on RE (-12.4\%) and MRC (-20.2\%) when evaluated on OOD test sets.
For both tasks, the OOD test set exhibits lower entity bias, achieving better performance on it suggests that the model relies less on entity bias as a predictive factor.

\textit{CoRE} and \textit{Continual Pretraining} are the only baselines that improve the ID performance. CoRE leads to a slight performance decrease on the OOD test set of RE in exchange,\footnote{This is because \textit{CoRE} is designed for a class-balanced setting, but this experiment emphasizes the performance on the raw class distribution. Moreover, we search its bias mitigation weight on the ID development set, which has a notably different entity distribution compared with the OOD test set.} while \textit{Continual Pretraining} further increases the OOD performance on MRC.
\textit{Entity Mask} successfully narrow down or even reverse the relative performance drop under OOD setting on the two tasks. However, its absolute performance decreases significantly due to the loss of predictive information from entities. Moreover, its effectiveness is dependent on the task property.
Unlike MRC, entities are given and are not answers in RE, so the gap between ID and OOD performance of \textit{Entity Mask} are much smaller.
\textit{Entity Substitution} stands out among all the baselines in terms of the OOD performance, with an absolute improvement of 3.5 points on RE and 7.7 points on MRC. However, its ID performance suffers a lot from the distribution shift of entities during training.

Our training-time intervention achieves the best OOD performance, with an absolute improvement of 2.2 points on RE and 1.4 points on MRC compared with \textit{Entity Substitution}. At the same time, its ID performance is also better. These results show that our method mitigates entity bias more effectively without losing much predictive information. In other words, the proposed method represents a better way to estimate the parameters of the proposed SCM accurately.

\begin{figure}[t]
\centering
\includegraphics[width=\columnwidth]{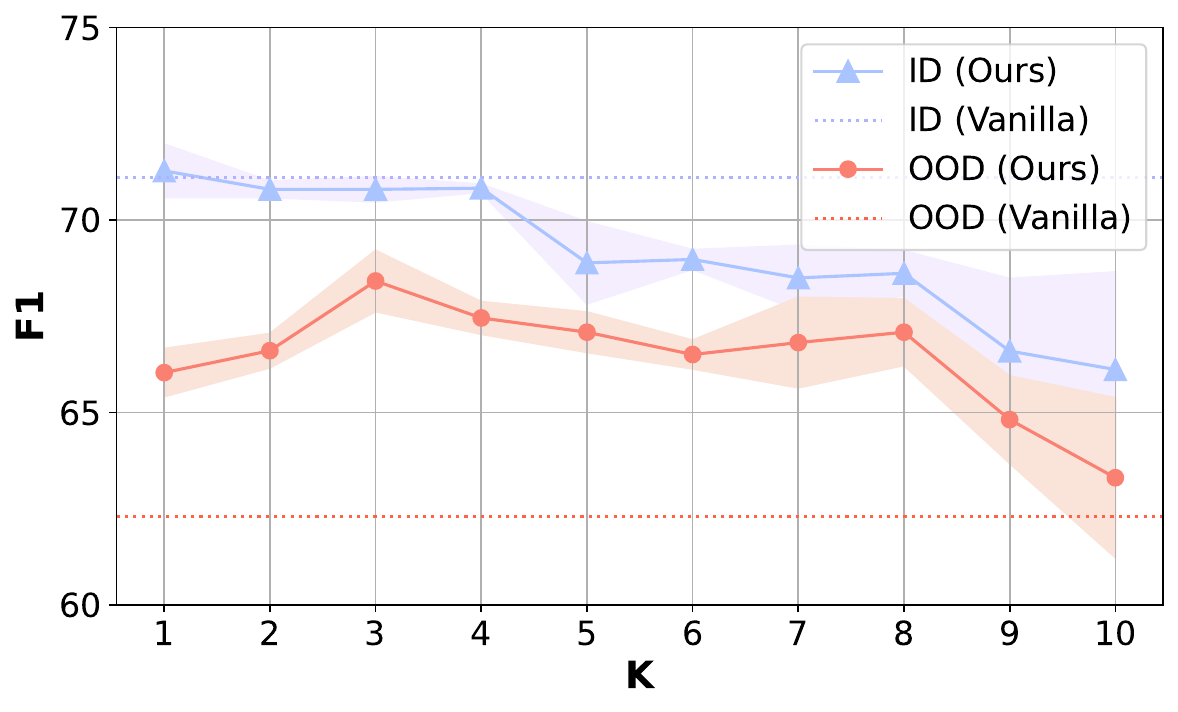}
\caption{F1 score of training-time intervention with different $k$ on RE.}
\label{fig/k}
\end{figure}

\subsection{Analysis}
\label{exp/white/analysis}
To provide a comprehensive understanding of our training-time intervention, we further conduct analyses on RE.

\stitle{Effect of \boldmath{$k$}}
The number of neighbors, $k$, plays a crucial role in balancing the predictive information and biasing information from entities. To find the sweet spot of $k$, we examine its influence on model performance as shown in \Cref{fig/k}. 
In general, the ID performance decreases when $k$ increases. As the value of $k$ increases, the resulting convex hull becomes larger, causing the center of the hull to move further away from the original entity. Consequently, both the predictive information and biasing information that contribute to ID performance gradually diminish.
In contrast, the OOD performance is lower when $k$ is too big or too small. When $k$ is too big, the same problem under ID setting also happens to the OOD setting. When $k$ is too small, the biasing information is not effectively mitigated, because the perturbed entity is too close to the original entity.

\stitle{Entity Type as Input}
Previous experiments in this section do not explicitly input entity information as it may disturb the causal analysis. 
Here, we analyze the effect of entity type information as input.
We use the \texttt{typed\_entity\_marker\_punct} input format from \citet{zhou2022improved}.
The ID and OOD F1 scores of vanilla RoBERTa model are 74.6 and 68.9 points, respectively.
Our training-time intervention further improves the ID performance by 0.7 points and the OOD performance by 2.9 points.
These results indicate that information from neighboring entities is complementary to coarse-grained entity type information for precise RE.

\begin{figure*}[t]
     \centering
     \begin{subfigure}[b]{0.245\textwidth}
         \centering
         \includegraphics[width=\textwidth]{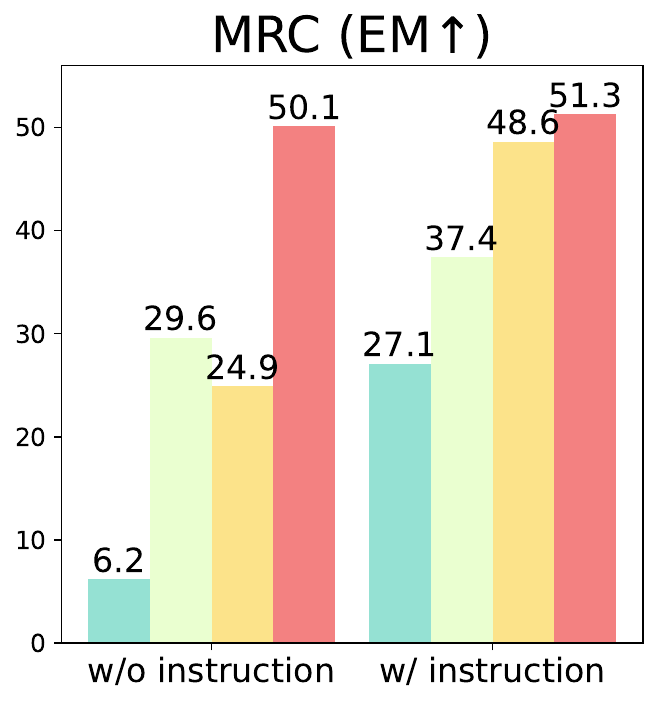}
         \label{gpt-1}
     \end{subfigure}
     \begin{subfigure}[b]{0.245\textwidth}
         \centering
         \includegraphics[width=\textwidth]{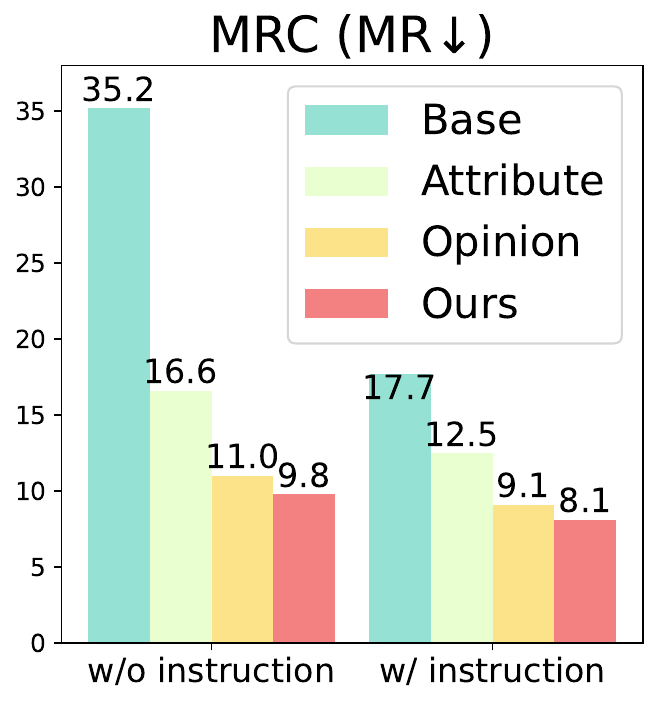}
         \label{gpt-1}
     \end{subfigure}
     \begin{subfigure}[b]{0.245\textwidth}
         \centering
         \includegraphics[width=\textwidth]{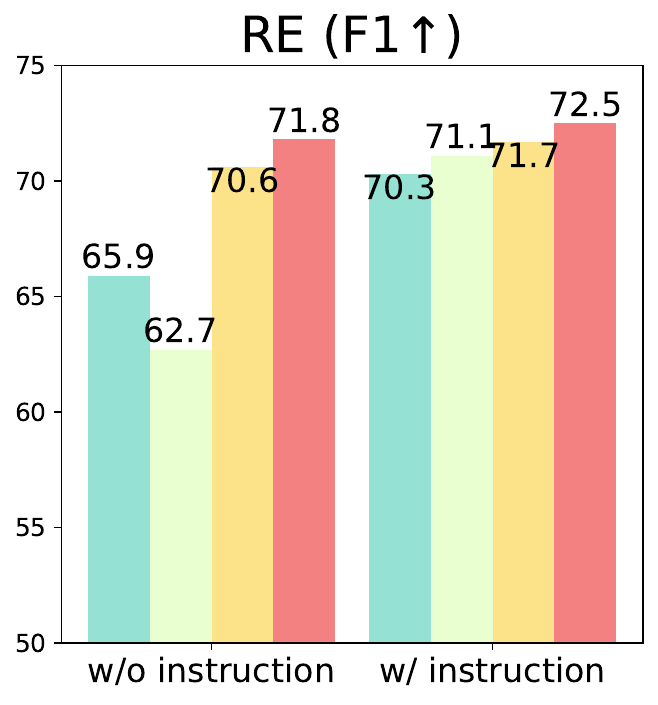}
         \label{gpt-1}
     \end{subfigure}
    \begin{subfigure}[b]{0.245\textwidth}
         \centering
         \includegraphics[width=\textwidth]{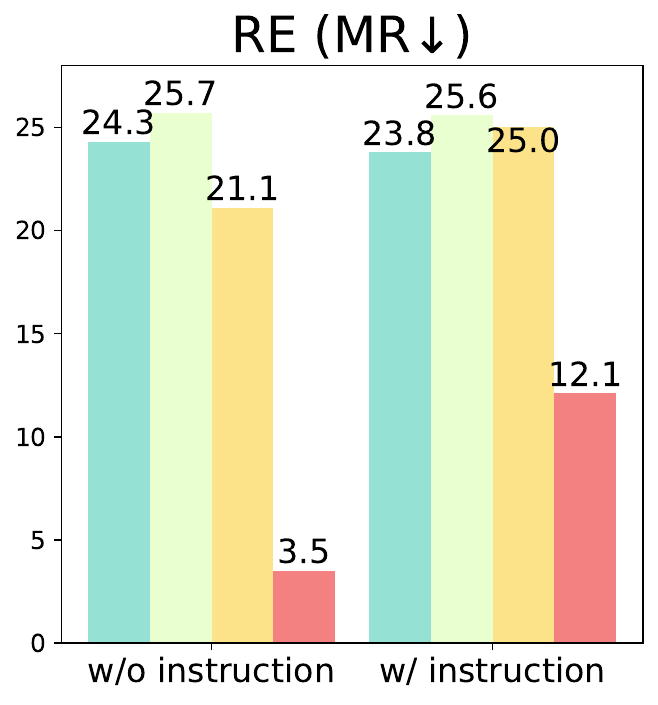}
         \label{gpt-1}
     \end{subfigure}
        \caption{GPT-3.5 results on MRC and RE under black-box setting. We report the EM score on MRC and the F1 score on RE, for which higher scores are better. We also report the MR score on both tasks, for which lower scores are better. Our in-context intervention performs consistently better than baselines under all settings.}
        \label{fig/gpt}
\end{figure*}

\begin{figure}[t]
\centering
\includegraphics[width=\columnwidth]{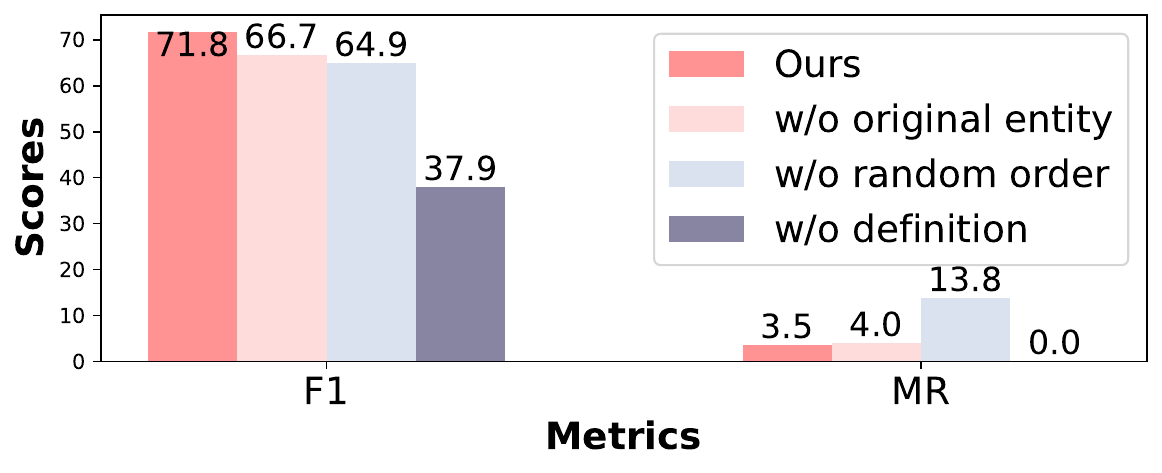}
\caption{Ablation study of in-context intervention for GPT-3.5 on RE.}
\label{fig/ablation}
\end{figure}

\section{Black-Box Experiments}
\label{sec/exp/black}

In this section, we evaluate our in-context intervention for mitigating entity bias from LLMs under black-box setting.

\subsection{Experimental Setup}

\stitle{Datasets}
Following \citet{zhou2023context}, we adopt GPT-3.5 \texttt{text-davinci-003} as the backbone LLM and evaluate the model performance under a zero-shot setting. 
We use the RE and MRC datasets provided by \citet{zhou2023context}. 
The RE dataset is based on Re-TACRED \cite{stoica2021re}. \citeauthor{zhou2023context} pair each instance's entities with a randomly sampled context that shares the same entity types but possesses different relations. To mitigate the influence of the label \textit{no\_relation}, which can also serve as a signal of abstention, we further filter out all instances whose original or updated labels are \textit{no\_relation}.
The MRC dataset is based on Natural Questions \cite{kwiatkowski2019natural}. \citeauthor{zhou2023context} replace the original answer in each instance with a randomly sampled entity of the same type.
They only collect instances where the LLM can give the correct answer based on the raw context.
Intuitively, LLMs that faithfully capture contextual information should update their answers based on the new context. %

\stitle{Metrics}
We report the F1 score for RE, and EM score for MRC.
To align with previous works, we also report the memorization ratio (MR; \citealt{longpre2021entity}) to measure the model's ability to update answers based on given contexts.\footnote{$MR = \frac{P_o}{P_o+P_s}$ , where $P_o$ is the probability that the model generates the original answer and $P_s$ is the probability that the model updates the answer correctly.}

\stitle{Baselines}
We compare our in-context intervention with the methods introduced by \citet{zhou2023context}.
\textit{Base} prompts directly concatenate the context and the question of each instance as the query.
\textit{Attribute}-based prompts append \textit{``in the given context''} to the question. 
\textit{Opinion}-based prompts modified the context to a narrator’s statement by prepending \textit{``Bob said''} to the context, and then query the LLM about the narrator’s opinion by prepending \textit{``What's Bob's opinion on''} to the question.
We evaluate all methods with and without specifically designed task instructions following \citet{zhou2023context}.

\stitle{Implementation Details}
We apply our in-context intervention to attribute-based prompts. 
We adopt the backbone LLM to propose two similar entities along with the original entity to define each placeholder.
To further eliminate the spurious entity mapping, we shuffle the entities for each placeholder before verbalization.
Details of all prompt templates used can be found in \Cref{sec/app/black}.
Since entities are not given in MRC, we detect named entities and replace them with placeholders using \texttt{gpt-3.5-turbo} as an external tool.
Given the potential abundance of entities in long contexts, we do not replace entities that exclusively appear in the context.

\subsection{Results}
As shown in \Cref{fig/gpt}, all methods benefit from carefully designed task instructions in terms of task performance. The \textit{Opinion}-based prompt performs the best among all baselines in most cases. Compared with the \textit{Base} prompt, it significantly improves the EM score by 18.7-21.5 points on MRC and the F1 score by 0.6-4.7 points on RE. Our in-context intervention achieves the highest EM/F1 score and the lowest MR score under all settings. Specifically, without task instruction, our in-context intervention outperforms the best baseline by 20.5 EM points on MRC and reduces the MR score by 17.6 points on RE. These results demonstrate the effectiveness of our causal intervention for addressing entity-based knowledge conflicts in black-box LLMs.

\subsection{Ablation Study}
\label{sec/black/ablation}

We in addition conduct an ablation study on RE to provide a comprehensive understanding of our method, as shown in \Cref{fig/ablation}.
When the placeholder definition is not provided (i.e., \textit{w/o definition}), no entity information, including both biasing and predictive information, appears in the input. As a result, it successfully blocks any spurious shortcuts with MR drops to 0. However, the F1 score also drops sharply from 71.8 points to 37.9 points, indicating that some entity information is essential to accurate RE and the LLM cannot understand the placeholders well without their definition.

We further examine the role of original entities in the placeholder definition. On the one hand, we remove the original entities from the definition (i.e., \textit{w/o original entity}). Results show that our method can still improve F1 while reducing MR. This verifies the effectiveness of using a set of similar entities to represent the predictive information from the original entity. On the other hand, we put the original subject and object entities at the same position (i.e., \textit{w/o entity shuffle}) in the definition so that the LLM can easily map them. As a result, the MR increases significantly, showing that the LLM can find spurious shortcuts even through mapping the subject entity and the object entity from two entity sets.

\section{Conclusion}

In this paper, we analyze the entity bias in LLMs from a causal view. 
Building upon an SCM whose parameters are easier to estimate, we propose training-time causal intervention for white-box LLMs and in-context causal intervention for black-box LLMs.
Both intervention techniques perturb the original entity with neighboring entities to mitigate spurious correlations between specific entities and predictions.
Experiments on relation extraction and machine reading comprehension show that the proposed intervention can effectively reduce the conflicts between parametric knowledge and contextual knowledge and significantly improve the performance of LLMs.
Future work can apply our causal intervention to more LLMs and tasks to achieve context-faithful answers. 

\section*{Acknowledgement}

We appreciate the reviewers for their insightful
comments and suggestions.
Fei Wang is supported by the Annenberg Fellowship and the Amazon ML Fellowship.
Wenjie Mo is supported by the USC CURVE Fellowship and the Provost's Research Fellowship.
Wenxuan Zhou and Muhao Chen are supported by the NSF Grant IIS 2105329, the NSF Grant ITE 2333736, 
the DARPA MCS program under Contract No. N660011924033 with
the United States Office Of Naval Research.
This work is also supported in part by a Cisco Research Award, two Amazon Research Awards, and a Keston Research Award.
Computing of this work has been partly supported by a subaward of NSF Cloudbank 1925001 through UCSD.

\section*{Limitation}
Although we have tried to verify the effectiveness of our method under diverse settings, including different LLMs, different accessibility of model parameters, and different tasks, %
there are always more options for further investigation, especially nowadays when more and more LLMs are kept produced.
Considering the property of the entity bias issue may vary when it comes to different LLMs and datasets from different domains, future work can build better benchmark for more comprehensive evaluation. 
In this paper, we only consider zero-shot prompting for black-box LLMs, because this will help us to control variables during causal analysis. However, it is possible to combine the proposed causal intervention with cutting-edge LLM inference methods, such as in-context learning \cite{brown2020language}, although the underlying SCM may become more complex.

\bibliography{reference}

\begin{thebibliography}{45}
\expandafter\ifx\csname natexlab\endcsname\relax\def\natexlab#1{#1}\fi

\bibitem[{Brown et~al.(2020)Brown, Mann, Ryder, Subbiah, Kaplan, Dhariwal,
  Neelakantan, Shyam, Sastry, Askell et~al.}]{brown2020language}
Tom Brown, Benjamin Mann, Nick Ryder, Melanie Subbiah, Jared~D Kaplan, Prafulla
  Dhariwal, Arvind Neelakantan, Pranav Shyam, Girish Sastry, Amanda Askell,
  et~al. 2020.
\newblock Language models are few-shot learners.
\newblock \emph{Advances in neural information processing systems},
  33:1877--1901.

\bibitem[{Chen et~al.(2022)Chen, Zhang, and Choi}]{chen-etal-2022-rich}
Hung-Ting Chen, Michael Zhang, and Eunsol Choi. 2022.
\newblock \href {https://aclanthology.org/2022.emnlp-main.146} {Rich knowledge
  sources bring complex knowledge conflicts: Recalibrating models to reflect
  conflicting evidence}.
\newblock In \emph{Proceedings of the 2022 Conference on Empirical Methods in
  Natural Language Processing}, pages 2292--2307, Abu Dhabi, United Arab
  Emirates. Association for Computational Linguistics.

\bibitem[{Dong et~al.(2021)Dong, Luu, Ji, and Liu}]{dong2021towards}
Xinshuai Dong, Anh~Tuan Luu, Rongrong Ji, and Hong Liu. 2021.
\newblock Towards robustness against natural language word substitutions.
\newblock In \emph{International Conference on Learning Representations}.

\bibitem[{Feder et~al.(2022)Feder, Keith, Manzoor, Pryzant, Sridhar,
  Wood-Doughty, Eisenstein, Grimmer, Reichart, Roberts
  et~al.}]{feder2022causal}
Amir Feder, Katherine~A Keith, Emaad Manzoor, Reid Pryzant, Dhanya Sridhar,
  Zach Wood-Doughty, Jacob Eisenstein, Justin Grimmer, Roi Reichart, Margaret~E
  Roberts, et~al. 2022.
\newblock Causal inference in natural language processing: Estimation,
  prediction, interpretation and beyond.
\newblock \emph{Transactions of the Association for Computational Linguistics},
  10:1138--1158.

\bibitem[{Finlayson et~al.(2021)Finlayson, Mueller, Gehrmann, Shieber, Linzen,
  and Belinkov}]{finlayson2021causal}
Matthew Finlayson, Aaron Mueller, Sebastian Gehrmann, Stuart Shieber, Tal
  Linzen, and Yonatan Belinkov. 2021.
\newblock Causal analysis of syntactic agreement mechanisms in neural language
  models.
\newblock \emph{arXiv preprint arXiv:2106.06087}.

\bibitem[{Hermann et~al.(2015)Hermann, Kocisky, Grefenstette, Espeholt, Kay,
  Suleyman, and Blunsom}]{hermann2015teaching}
Karl~Moritz Hermann, Tomas Kocisky, Edward Grefenstette, Lasse Espeholt, Will
  Kay, Mustafa Suleyman, and Phil Blunsom. 2015.
\newblock Teaching machines to read and comprehend.
\newblock \emph{Advances in neural information processing systems}, 28.

\bibitem[{Jiang et~al.(2020)Jiang, Xu, Araki, and
  Neubig}]{jiang-etal-2020-know}
Zhengbao Jiang, Frank~F. Xu, Jun Araki, and Graham Neubig. 2020.
\newblock \href {https://doi.org/10.1162/tacl_a_00324} {How can we know what
  language models know?}
\newblock \emph{Transactions of the Association for Computational Linguistics},
  8:423--438.

\bibitem[{Joshi et~al.(2020)Joshi, Chen, Liu, Weld, Zettlemoyer, and
  Levy}]{joshi2020spanbert}
Mandar Joshi, Danqi Chen, Yinhan Liu, Daniel~S Weld, Luke Zettlemoyer, and Omer
  Levy. 2020.
\newblock Spanbert: Improving pre-training by representing and predicting
  spans.
\newblock \emph{Transactions of the Association for Computational Linguistics},
  8:64--77.

\bibitem[{Joshi et~al.(2017)Joshi, Choi, Weld, and
  Zettlemoyer}]{joshi2017triviaqa}
Mandar Joshi, Eunsol Choi, Daniel~S Weld, and Luke Zettlemoyer. 2017.
\newblock Triviaqa: A large scale distantly supervised challenge dataset for
  reading comprehension.
\newblock In \emph{Proceedings of the 55th Annual Meeting of the Association
  for Computational Linguistics (Volume 1: Long Papers)}, pages 1601--1611.

\bibitem[{Kasai et~al.(2022)Kasai, Sakaguchi, Takahashi, Bras, Asai, Yu, Radev,
  Smith, Choi, and Inui}]{kasai2022realtime}
Jungo Kasai, Keisuke Sakaguchi, Yoichi Takahashi, Ronan~Le Bras, Akari Asai,
  Xinyan Yu, Dragomir Radev, Noah~A Smith, Yejin Choi, and Kentaro Inui. 2022.
\newblock Realtime qa: What's the answer right now?
\newblock \emph{arXiv preprint arXiv:2207.13332}.

\bibitem[{Keith et~al.(2020)Keith, Jensen, and O’Connor}]{keith2020text}
Katherine Keith, David Jensen, and Brendan O’Connor. 2020.
\newblock Text and causal inference: A review of using text to remove
  confounding from causal estimates.
\newblock In \emph{Proceedings of the 58th Annual Meeting of the Association
  for Computational Linguistics}, pages 5332--5344.

\bibitem[{Kwiatkowski et~al.(2019)Kwiatkowski, Palomaki, Redfield, Collins,
  Parikh, Alberti, Epstein, Polosukhin, Devlin, Lee
  et~al.}]{kwiatkowski2019natural}
Tom Kwiatkowski, Jennimaria Palomaki, Olivia Redfield, Michael Collins, Ankur
  Parikh, Chris Alberti, Danielle Epstein, Illia Polosukhin, Jacob Devlin,
  Kenton Lee, et~al. 2019.
\newblock Natural questions: a benchmark for question answering research.
\newblock \emph{Transactions of the Association for Computational Linguistics},
  7:453--466.

\bibitem[{Lalor et~al.(2022)Lalor, Yang, Smith, Forsgren, and
  Abbasi}]{lalor2022benchmarking}
John~P Lalor, Yi~Yang, Kendall Smith, Nicole Forsgren, and Ahmed Abbasi. 2022.
\newblock Benchmarking intersectional biases in nlp.
\newblock In \emph{Proceedings of the 2022 Conference of the North American
  Chapter of the Association for Computational Linguistics: Human Language
  Technologies}, pages 3598--3609.

\bibitem[{Li et~al.(2022)Li, Rawat, Zaheer, Wang, Lukasik, Veit, Yu, and
  Kumar}]{li2022large}
Daliang Li, Ankit~Singh Rawat, Manzil Zaheer, Xin Wang, Michal Lukasik, Andreas
  Veit, Felix Yu, and Sanjiv Kumar. 2022.
\newblock Large language models with controllable working memory.
\newblock \emph{arXiv preprint arXiv:2211.05110}.

\bibitem[{Lin et~al.(2022)Lin, Hilton, and Evans}]{lin-etal-2022-truthfulqa}
Stephanie Lin, Jacob Hilton, and Owain Evans. 2022.
\newblock \href {https://doi.org/10.18653/v1/2022.acl-long.229}
  {{T}ruthful{QA}: Measuring how models mimic human falsehoods}.
\newblock In \emph{Proceedings of the 60th Annual Meeting of the Association
  for Computational Linguistics (Volume 1: Long Papers)}, pages 3214--3252,
  Dublin, Ireland. Association for Computational Linguistics.

\bibitem[{Liska et~al.(2022)Liska, Kocisky, Gribovskaya, Terzi, Sezener,
  Agrawal, Cyprien De~Masson, Scholtes, Zaheer, Young
  et~al.}]{liska2022streamingqa}
Adam Liska, Tomas Kocisky, Elena Gribovskaya, Tayfun Terzi, Eren Sezener,
  Devang Agrawal, D’Autume Cyprien De~Masson, Tim Scholtes, Manzil Zaheer,
  Susannah Young, et~al. 2022.
\newblock Streamingqa: A benchmark for adaptation to new knowledge over time in
  question answering models.
\newblock In \emph{International Conference on Machine Learning}, pages
  13604--13622. PMLR.

\bibitem[{Liu et~al.(2019)Liu, Ott, Goyal, Du, Joshi, Chen, Levy, Lewis,
  Zettlemoyer, and Stoyanov}]{liu2019roberta}
Yinhan Liu, Myle Ott, Naman Goyal, Jingfei Du, Mandar Joshi, Danqi Chen, Omer
  Levy, Mike Lewis, Luke Zettlemoyer, and Veselin Stoyanov. 2019.
\newblock Roberta: A robustly optimized bert pretraining approach.
\newblock \emph{arXiv preprint arXiv:1907.11692}.

\bibitem[{Longpre et~al.(2021)Longpre, Perisetla, Chen, Ramesh, DuBois, and
  Singh}]{longpre2021entity}
Shayne Longpre, Kartik Perisetla, Anthony Chen, Nikhil Ramesh, Chris DuBois,
  and Sameer Singh. 2021.
\newblock Entity-based knowledge conflicts in question answering.
\newblock In \emph{Proceedings of the 2021 Conference on Empirical Methods in
  Natural Language Processing}, pages 7052--7063.

\bibitem[{Nan et~al.(2021)Nan, Zeng, Qiao, Guo, and Lu}]{nan2021uncovering}
Guoshun Nan, Jiaqi Zeng, Rui Qiao, Zhijiang Guo, and Wei Lu. 2021.
\newblock Uncovering main causalities for long-tailed information extraction.
\newblock In \emph{Proceedings of the 2021 Conference on Empirical Methods in
  Natural Language Processing}, pages 9683--9695.

\bibitem[{Niu and Zhang(2021)}]{niu2021introspective}
Yulei Niu and Hanwang Zhang. 2021.
\newblock Introspective distillation for robust question answering.
\newblock \emph{Advances in Neural Information Processing Systems},
  34:16292--16304.

\bibitem[{Pearl(2012)}]{pearl2012calculus}
Judea Pearl. 2012.
\newblock The do-calculus revisited.
\newblock In \emph{Proceedings of the Twenty-Eighth Conference on Uncertainty
  in Artificial Intelligence}, pages 3--11.

\bibitem[{Peng et~al.(2020)Peng, Gao, Han, Lin, Li, Liu, Sun, and
  Zhou}]{peng2020learning}
Hao Peng, Tianyu Gao, Xu~Han, Yankai Lin, Peng Li, Zhiyuan Liu, Maosong Sun,
  and Jie Zhou. 2020.
\newblock Learning from context or names? an empirical study on neural relation
  extraction.
\newblock In \emph{Proceedings of the 2020 Conference on Empirical Methods in
  Natural Language Processing (EMNLP)}, pages 3661--3672.

\bibitem[{Petroni et~al.(2019)Petroni, Rockt{\"a}schel, Riedel, Lewis, Bakhtin,
  Wu, and Miller}]{petroni-etal-2019-language}
Fabio Petroni, Tim Rockt{\"a}schel, Sebastian Riedel, Patrick Lewis, Anton
  Bakhtin, Yuxiang Wu, and Alexander Miller. 2019.
\newblock \href {https://doi.org/10.18653/v1/D19-1250} {Language models as
  knowledge bases?}
\newblock In \emph{Proceedings of the 2019 Conference on Empirical Methods in
  Natural Language Processing and the 9th International Joint Conference on
  Natural Language Processing (EMNLP-IJCNLP)}, pages 2463--2473, Hong Kong,
  China. Association for Computational Linguistics.

\bibitem[{Qian et~al.(2021{\natexlab{a}})Qian, Feng, Wen, Ma, and
  Xie}]{qian2021counterfactual}
Chen Qian, Fuli Feng, Lijie Wen, Chunping Ma, and Pengjun Xie.
  2021{\natexlab{a}}.
\newblock Counterfactual inference for text classification debiasing.
\newblock In \emph{Proceedings of the 59th Annual Meeting of the Association
  for Computational Linguistics and the 11th International Joint Conference on
  Natural Language Processing (Volume 1: Long Papers)}, pages 5434--5445.

\bibitem[{Qian et~al.(2021{\natexlab{b}})Qian, Beirami, Lin, De, Geramifard,
  Yu, and Sankar}]{qian-etal-2021-annotation}
Kun Qian, Ahmad Beirami, Zhouhan Lin, Ankita De, Alborz Geramifard, Zhou Yu,
  and Chinnadhurai Sankar. 2021{\natexlab{b}}.
\newblock \href {https://aclanthology.org/2021.sigdial-1.35} {Annotation
  inconsistency and entity bias in {M}ulti{WOZ}}.
\newblock In \emph{Proceedings of the 22nd Annual Meeting of the Special
  Interest Group on Discourse and Dialogue}, pages 326--337, Singapore and
  Online. Association for Computational Linguistics.

\bibitem[{Roberts et~al.(2020)Roberts, Raffel, and
  Shazeer}]{roberts-etal-2020-much}
Adam Roberts, Colin Raffel, and Noam Shazeer. 2020.
\newblock \href {https://doi.org/10.18653/v1/2020.emnlp-main.437} {How much
  knowledge can you pack into the parameters of a language model?}
\newblock In \emph{Proceedings of the 2020 Conference on Empirical Methods in
  Natural Language Processing (EMNLP)}, pages 5418--5426, Online. Association
  for Computational Linguistics.

\bibitem[{Stoica et~al.(2021)Stoica, Platanios, and P{\'o}czos}]{stoica2021re}
George Stoica, Emmanouil~Antonios Platanios, and Barnab{\'a}s P{\'o}czos. 2021.
\newblock Re-tacred: Addressing shortcomings of the tacred dataset.
\newblock In \emph{Proceedings of the AAAI Conference on Artificial
  Intelligence}, volume~35, pages 13843--13850.

\bibitem[{Sweeney and Najafian(2019)}]{sweeney2019transparent}
Chris Sweeney and Maryam Najafian. 2019.
\newblock A transparent framework for evaluating unintended demographic bias in
  word embeddings.
\newblock In \emph{Proceedings of the 57th Annual Meeting of the Association
  for Computational Linguistics}, pages 1662--1667.

\bibitem[{Tian et~al.(2022)Tian, Cao, Zhang, and Xing}]{tian2022debiasing}
Bing Tian, Yixin Cao, Yong Zhang, and Chunxiao Xing. 2022.
\newblock Debiasing nlu models via causal intervention and counterfactual
  reasoning.
\newblock In \emph{Proceedings of the AAAI Conference on Artificial
  Intelligence}, volume~36, pages 11376--11384.

\bibitem[{Udomcharoenchaikit et~al.(2022)Udomcharoenchaikit, Ponwitayarat,
  Payoungkhamdee, Masuk, Buaphet, Chuangsuwanich, and
  Nutanong}]{udomcharoenchaikit2022mitigating}
Can Udomcharoenchaikit, Wuttikorn Ponwitayarat, Patomporn Payoungkhamdee,
  Kanruethai Masuk, Weerayut Buaphet, Ekapol Chuangsuwanich, and Sarana
  Nutanong. 2022.
\newblock Mitigating spurious correlation in natural language understanding
  with counterfactual inference.
\newblock In \emph{Proceedings of the 2022 Conference on Empirical Methods in
  Natural Language Processing}, pages 11308--11321.

\bibitem[{Venkit and Wilson(2021)}]{venkit2021identification}
Pranav~Narayanan Venkit and Shomir Wilson. 2021.
\newblock Identification of bias against people with disabilities in sentiment
  analysis and toxicity detection models.
\newblock \emph{arXiv preprint arXiv:2111.13259}.

\bibitem[{Verma and Pearl(1990)}]{verma1990equivalence}
Thomas Verma and Judea Pearl. 1990.
\newblock Equivalence and synthesis of causal models.
\newblock In \emph{Proceedings of the Sixth Annual Conference on Uncertainty in
  Artificial Intelligence}, pages 255--270.

\bibitem[{Vig et~al.(2020)Vig, Gehrmann, Belinkov, Qian, Nevo, Sakenis, Huang,
  Singer, and Shieber}]{vig2020causal}
Jesse Vig, Sebastian Gehrmann, Yonatan Belinkov, Sharon Qian, Daniel Nevo,
  Simas Sakenis, Jason Huang, Yaron Singer, and Stuart Shieber. 2020.
\newblock Causal mediation analysis for interpreting neural nlp: The case of
  gender bias.
\newblock \emph{arXiv preprint arXiv:2004.12265}.

\bibitem[{Wang et~al.(2022)Wang, Chen, Zhou, Cai, Liang, Liu, Yang, Liu, and
  Hooi}]{wang2022should}
Yiwei Wang, Muhao Chen, Wenxuan Zhou, Yujun Cai, Yuxuan Liang, Dayiheng Liu,
  Baosong Yang, Juncheng Liu, and Bryan Hooi. 2022.
\newblock Should we rely on entity mentions for relation extraction? debiasing
  relation extraction with counterfactual analysis.
\newblock \emph{arXiv preprint arXiv:2205.03784}.

\bibitem[{Wang et~al.(2023)Wang, Hooi, Wang, Cai, Liang, Zhou, Tang, Duan, and
  Chen}]{wang2023entred}
Yiwei Wang, Bryan Hooi, Fei Wang, Yujun Cai, Yuxuan Liang, Wenxuan Zhou, Jing
  Tang, Manjuan Duan, and Muhao Chen. 2023.
\newblock How fragile is relation extraction under entity replacements?
\newblock In \emph{Proceedings of the 27th SIGNLL Conference on Computational
  Natural Language Learning (CoNLL)}.

\bibitem[{Wei et~al.(2022)Wei, Bosma, Zhao, Guu, Yu, Lester, Du, Dai, and
  Le}]{weifinetuned}
Jason Wei, Maarten Bosma, Vincent Zhao, Kelvin Guu, Adams~Wei Yu, Brian Lester,
  Nan Du, Andrew~M Dai, and Quoc~V Le. 2022.
\newblock Finetuned language models are zero-shot learners.
\newblock In \emph{International Conference on Learning Representations}.

\bibitem[{Weld et~al.(2022)Weld, West, Glenski, Arbour, Rossi, and
  Althoff}]{weld2022adjusting}
Galen Weld, Peter West, Maria Glenski, David Arbour, Ryan~A Rossi, and Tim
  Althoff. 2022.
\newblock Adjusting for confounders with text: Challenges and an empirical
  evaluation framework for causal inference.
\newblock In \emph{Proceedings of the International AAAI Conference on Web and
  Social Media}, volume~16, pages 1109--1120.

\bibitem[{Xu et~al.(2022)Xu, Wang, Li, Dong, and Chen}]{xu2022does}
Nan Xu, Fei Wang, Bangzheng Li, Mingtao Dong, and Muhao Chen. 2022.
\newblock Does your model classify entities reasonably? diagnosing and
  mitigating spurious correlations in entity typing.
\newblock In \emph{Proceedings of the 2022 Conference on Empirical Methods in
  Natural Language Processing}.

\bibitem[{Yan et~al.(2022)Yan, Xiao, Mukherjee, Lin, Jia, and
  Ren}]{yan2022robustness}
Jun Yan, Yang Xiao, Sagnik Mukherjee, Bill~Yuchen Lin, Robin Jia, and Xiang
  Ren. 2022.
\newblock On the robustness of reading comprehension models to entity renaming.
\newblock In \emph{Proceedings of the 2022 Conference of the North American
  Chapter of the Association for Computational Linguistics: Human Language
  Technologies}, pages 508--520.

\bibitem[{Zhang et~al.(2020{\natexlab{a}})Zhang, Zhang, Tang, Hua, and
  Sun}]{zhang2020causal}
Dong Zhang, Hanwang Zhang, Jinhui Tang, Xian-Sheng Hua, and Qianru Sun.
  2020{\natexlab{a}}.
\newblock Causal intervention for weakly-supervised semantic segmentation.
\newblock \emph{Advances in Neural Information Processing Systems},
  33:655--666.

\bibitem[{Zhang et~al.(2020{\natexlab{b}})Zhang, Bai, Zhang, Bai, Zhu, and
  Zhao}]{zhang2020demographics}
Guanhua Zhang, Bing Bai, Junqi Zhang, Kun Bai, Conghui Zhu, and Tiejun Zhao.
  2020{\natexlab{b}}.
\newblock Demographics should not be the reason of toxicity: Mitigating
  discrimination in text classifications with instance weighting.
\newblock \emph{arXiv preprint arXiv:2004.14088}.

\bibitem[{Zhang et~al.(2017)Zhang, Zhong, Chen, Angeli, and
  Manning}]{zhang2017position}
Yuhao Zhang, Victor Zhong, Danqi Chen, Gabor Angeli, and Christopher~D Manning.
  2017.
\newblock Position-aware attention and supervised data improve slot filling.
\newblock In \emph{Proceedings of the 2017 Conference on Empirical Methods in
  Natural Language Processing}, pages 35--45.

\bibitem[{Zhou and Chen(2022)}]{zhou2022improved}
Wenxuan Zhou and Muhao Chen. 2022.
\newblock An improved baseline for sentence-level relation extraction.
\newblock In \emph{Proceedings of the 2nd Conference of the Asia-Pacific
  Chapter of the Association for Computational Linguistics and the 12th
  International Joint Conference on Natural Language Processing}, pages
  161--168.

\bibitem[{Zhou et~al.(2023)Zhou, Zhang, Poon, and Chen}]{zhou2023context}
Wenxuan Zhou, Sheng Zhang, Hoifung Poon, and Muhao Chen. 2023.
\newblock Context-faithful prompting for large language models.
\newblock In \emph{Findings of the 2023 Conference on Empirical Methods in
  Natural Language Processing}.

\bibitem[{Zhu et~al.(2022)Zhu, Sheng, Cao, Li, Wang, and
  Zhuang}]{zhu2022generalizing}
Yongchun Zhu, Qiang Sheng, Juan Cao, Shuokai Li, Danding Wang, and Fuzhen
  Zhuang. 2022.
\newblock Generalizing to the future: Mitigating entity bias in fake news
  detection.
\newblock In \emph{Proceedings of the 45th International ACM SIGIR Conference
  on Research and Development in Information Retrieval}, pages 2120--2125.

\end{thebibliography}
\bibliographystyle{acl_natbib}

\clearpage

\appendix

\section{Implementation Details}

\subsection{White-Box Experiments}
\label{sec/app/white}

For RE, we use RoBERTa-Large as our backbone model, which has 354 million parameters.
Our implementation is based on the codebase by \citet{zhou2022improved} with their default hyper-parameters. 
More specifically, we employ a learning rate of 3e-5, a batch size of 32, and conduct training for a total of 5 epochs. 
Other method-specific hyper-parameters are selected on the development set of TACRED. 
Finetuning typically takes 1.5 hours on an NVIDIA RTX A5000 GPU.

For MRC, we use SpanBERT-base-cased as our backbone model, which has 110 million parameters.
Our implementation is based on the codebase by \citet{yan2022robustness} with their default hyper-parameters.
More specifically, we employ a learning rate of 2e-5, a batch size of 16, and conduct training for a total of 4 epochs. 
Other method-specific hyper-parameters are selected on the hold-out development set of TriviaQA.
Finetuning typically takes 3 hours on an NVIDIA RTX A5000 GPU.

\subsection{Black-Box Experiments}
\label{sec/app/black}

\definecolor{green1}{RGB}{0, 176, 80}
\definecolor{blue1}{RGB}{46, 117, 182}
\definecolor{orange1}{RGB}{197, 90, 17}
\definecolor{instruct_blue}{RGB}{217, 241, 253}
\definecolor{instruct_green}{RGB}{211, 245, 199}
\definecolor{instruct_yellow}{RGB}{255, 242, 204}
\definecolor{bd_blue}{RGB}{68, 114, 196}

Our implementation is based on the codebase by \citet{zhou2023context}.

The instruction for MRC is 
\begin{tcolorbox}[colback=instruct_yellow,colframe=bd_blue,boxsep=1pt,left=9pt,right=9pt,top=6pt,bottom=6pt,boxrule=1pt]
Instruction: read the given information and answer the corresponding question.
\end{tcolorbox}

The prompt without instruction for MRC is 

\begin{tcolorbox}[colback=instruct_yellow,colframe=bd_blue,boxsep=1pt,left=9pt,right=9pt,top=6pt,bottom=6pt,boxrule=1pt]
Assume that \{ENTITY0\} can be any of \{entity0\_candidates\}. [Assume that \{ENTITY1\} can be any of \{entity1\_candidates\} ...]

\{context\}

Q:\{question\} based on the given text? Extract the answer from the given text. Do not add other words.

A:

\end{tcolorbox}

The instruction for RE is 
\begin{tcolorbox}[colback=instruct_yellow,colframe=bd_blue,boxsep=1pt,left=9pt,right=9pt,top=6pt,bottom=6pt,boxrule=1pt]
Identify the relationship between two entities from a list of options.
\end{tcolorbox}

The prompt without instruction for RE is 

\begin{tcolorbox}[colback=instruct_yellow,colframe=bd_blue,boxsep=1pt,left=9pt,right=9pt,top=6pt,bottom=6pt,boxrule=1pt]
Assume that subject\_entity is one of \{subj\_candidates\}, while object\_entity is one of \{obj\_candidates\} in the following text.

\{context\}

Q: Which option indicates the relationship between subject\_entity and object\_entity in the given text?

Options:\{options\}

A:
\end{tcolorbox}

The prompt template for detecting entities in MRC is 
\begin{tcolorbox}[colback=instruct_yellow,colframe=bd_blue,boxsep=1pt,left=9pt,right=9pt,top=6pt,bottom=6pt,boxrule=1pt]
List named entities in the following sentence. Separate the entities with \#\#\#, if you find multiple entities. Do not add additional words before or after your answers.\\ \{sentence\}
\end{tcolorbox}

The prompt template for replacing entities with placeholders in MRC is 
\begin{tcolorbox}[colback=instruct_yellow,colframe=bd_blue,boxsep=1pt,left=9pt,right=9pt,top=6pt,bottom=6pt,boxrule=1pt]
Replace the entity \{entity\_list\} in the following paragraph. \\ \{paragraph\} 
\end{tcolorbox}

The prompt template for finding similar entities is 
\begin{tcolorbox}[colback=instruct_yellow,colframe=bd_blue,boxsep=1pt,left=9pt,right=9pt,top=6pt,bottom=6pt,boxrule=1pt]
Name two [\{entity\_type\}] entities similar to "\{entity\}". Separate the entities with \#\#\#, and do not add additional words before or after your answers. Provide random answers if you are not sure.
\end{tcolorbox}

In all the above prompts, variables are surrounded with curly brackets and optional variables are surrounded with square brackets.

\end{document}